\documentclass[10pt,twocolumn,letterpaper]{article}

\usepackage[pagenumbers]{cvpr} %

\makeatletter
\@namedef{ver@everyshi.sty}{}
\makeatother
\usepackage{tikz}

\usepackage{graphicx}
\usepackage{amsmath}
\usepackage{amssymb}
\usepackage{booktabs}

\usepackage[hang,flushmargin]{footmisc}

\usepackage{algorithm, algpseudocode}  %

\usepackage[toc,page]{appendix} 
\usepackage{pifont}%
\usepackage{xcolor}
\usepackage{xspace}
\usepackage{multirow}
\usepackage{arydshln}

\usepackage{tabularray}
\UseTblrLibrary{booktabs} 

\definecolor{noel_color}{RGB}{10, 114, 72}
\definecolor{noel_green}{RGB}{61,137,37}
\definecolor{noel_green_v2}{RGB}{79,126,127}
\definecolor{noelgray}{RGB}{50,50,50}

\definecolor{semanticcolour}{RGB}{210,226,241}
\definecolor{instancecolour}{RGB}{239,255,253}
\definecolor{maskproposalscolour}{RGB}{248,229,208}

\newcommand{\comment}[1]{{\textcolor{gray}{\# #1}}}

\newcommand{\xmark}{\ding{55}}%

\newcommand{\mypara}[1]{\vspace{2pt}\noindent{\bf{#1}}}
\newcommand{\methodName}{\textsc{ZUTIS}\xspace}

\newcommand{\mcal}[1]{\mathcal{#1}}
\newcommand{\mbb}[1]{\mathbb{#1}}
\newcommand{\mbf}[1]{\mathbf{#1}}

\newcommand{\green}[1]{\textcolor{noel_green}{#1}}
\newcommand{\commentgreen}[1]{\textcolor{noel_green_v2}{#1}}

\newcommand{\maskap}[0]{\text{AP}^{\text{mk}}}
\newcommand{\maskapfifty}[0]{\text{AP}^{\text{mk}}_{50}}
\newcommand{\maskapseventyfive}[0]{\text{AP}^{\text{mk}}_{75}}

\newcommand{\qcr}[1]{{\fontfamily{qcr}\selectfont{#1}}}

\newcommand{\imageEncoder}{$\phi_\mcal{I}$\xspace}
\newcommand{\textEncoder}{$\phi_\mcal{T}$\xspace}
\newcommand{\indexDataset}{$\mcal{U}$\xspace}

\makeatletter
\def\thickhline{%
  \noalign{\ifnum0=`}\fi\hrule \@height \thickarrayrulewidth \futurelet
   \reserved@a\@xthickhline}
\def\@xthickhline{\ifx\reserved@a\thickhline
               \vskip\doublerulesep
               \vskip-\thickarrayrulewidth
             \fi
      \ifnum0=`{\fi}}
\makeatother

\newlength{\thickarrayrulewidth}
\renewcommand{\comment}[1]{\commentgreen{\# #1}}

\usepackage{tikz}
\usetikzlibrary{fit,calc}
\newcommand*{\tikzmk}[1]{\tikz[remember picture,overlay] \node (#1) {};\ignorespaces}
\newcommand{\boxit}[1]{\tikz[remember picture,overlay]{\node[yshift=3pt,fill=#1,fill opacity=.23, fit={(A)($(B)+(.95\linewidth,.8\baselineskip)$)}] {};}\ignorespaces}
\colorlet{pink}{red!40}
\colorlet{blue}{cyan!60}

\usepackage[pagebackref,breaklinks,colorlinks]{hyperref}

\usepackage[capitalize]{cleveref}
\crefname{section}{Sec.}{Secs.}
\Crefname{section}{Section}{Sections}
\Crefname{table}{Table}{Tables}
\crefname{table}{Tab.}{Tabs.}

\def\cvprPaperID{10163} %
\def\confName{CVPR}
\def\confYear{2023}

\begin{document}

\title{Zero-shot Unsupervised Transfer Instance Segmentation}

\author{Gyungin Shin$^{1,2}$ \quad \quad \quad Samuel Albanie$^2$ \quad \quad \quad Weidi Xie$^{1,3}$\\ [2pt]
$^1$Visual Geometry Group,  University of Oxford, UK\\
$^2$Cambridge Applied Machine Learning Lab,  University of Cambridge, UK\\
$^3$Cooperative Medianet Innovation Center, Shanghai Jiao Tong University, China\\
{\tt\small \url{https://www.robots.ox.ac.uk/~vgg/research/zutis}
}}
\maketitle

\begin{abstract}
Segmentation is a core computer vision competency, with applications spanning a broad range of scientifically and economically valuable domains.
To date, however, the prohibitive cost of annotation has limited the deployment of flexible segmentation models.
In this work, we propose \textbf{Z}ero-shot \textbf{U}nsupervised \textbf{T}ransfer \textbf{I}nstance \textbf{S}egmentation (\methodName), 
a framework that aims to meet this challenge.
The key strengths of \methodName are:
(i) no requirement for instance-level or pixel-level annotations;
(ii) an ability of zero-shot transfer, i.e., no assumption on access to a target data distribution;
(iii) a unified framework for semantic and instance segmentations with solid performance on both tasks compared to state-of-the-art unsupervised methods.
While comparing to previous work, 
we show \methodName achieves a gain of \textbf{2.2 mask AP} on COCO-20K and \textbf{14.5 mIoU} on ImageNet-S with 919 categories for instance and semantic segmentations, respectively.
The code is made publicly available.\footnote{\url{https://github.com/NoelShin/zutis}}
\vspace{-5mm}
\end{abstract}

\section{Introduction}
\vspace{-2mm}
In computer vision, the task of segmentation aims to group pixels within an image into coherent, meaningful regions.
Accurate segmentation unlocks a host of applications such as tumour assessment in medical images~\cite{bilic2019liver}, land cover estimation~\cite{tong2020land} for logistical planning, scene segmentation for autonomous driving~\cite{cordts2016cvpr}, to name a few.
The central challenge that limits the deployment of such applications is the high cost of obtaining large, accurate collections of pixel-level annotations to train appropriate segmenters. For example, it was reported that when constructing the Cityscapes dataset, it took 90 minutes to fully annotate and validate individual images~\cite{cordts2016cvpr}.

\begin{figure}[!t]
    \centering
    \includegraphics[width=0.48\textwidth]{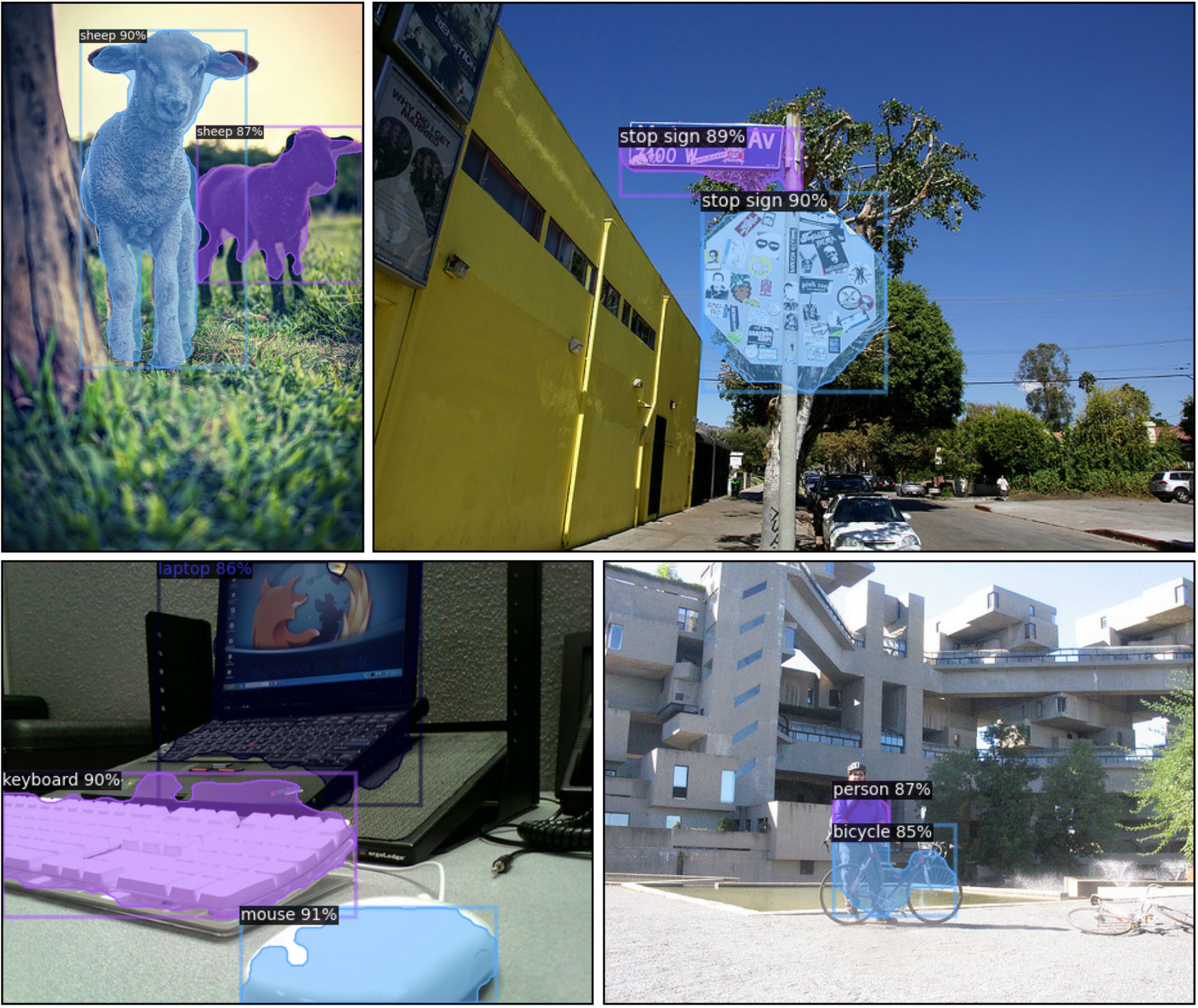}
    \caption{We propose \methodName, a framework for zero-shot unsupervised transfer instance segmentation.
    The figure depicts instance segmentations made by \methodName on COCO-20K~\cite{vo2022eccv} and VOC2012~\cite{Everingham10}. 
    Without pixel-level annotation or access to the target distribution, \methodName acquires the ability to reliably segment instances within an image.}
    \vspace{-5mm}
    \label{fig:teaser}
\end{figure}

To overcome this challenge, a range of unsupervised segmentation methods have been developed that forgo pixel-level supervision~\cite{ji2019iccv,cho2021cvpr,gansbeke2022maskdistill,zhang2020self,hamilton2022iclr,vobecky2022arxiv}.
One particularly promising line of work has focused on a setting known as \textit{unsupervised semantic segmentation with language-image pretraining} (USSLIP)~\cite{shin2022reco,shin2022namedmask}, which leverages a vision-language foundation model~\cite{bommasani2021opportunities} that has been pretrained on a large corpus of internet-sourced image-text pairs. USSLIP methods exhibit strong segmentation performance, category label flexibility and zero-shot transfer---the ability to perform well on a downstream task without access to images from the target distribution.
However, while USSLIP methods enable \textit{semantic segmentation}, 
no such method developed to date possesses the ability to differentiate between \textit{instances within a semantic category}, a key functionality for many fine-grained applications.

In this paper, we consider a challenging task, 
\textbf{Z}ero-shot \textbf{U}nsupervised \textbf{T}ransfer \textbf{I}nstance \textbf{S}egmentation, {\em i.e.}, to segment the instances present in an image and infer its semantic classes without relying on manual supervision or access to a target dataset.
To tackle such challenge, we start from the recent progress in USSLIP~\cite{shin2022namedmask}, retrieving images for given concept with a pretrained visual language model (e.g., CLIP), then generating pseudo-masks for the collected images with an unsupervised saliency detector.
To take one step further, we extend the prior USSLIP architectures with two critical abilities, namely, instance-level segmentation and generalisation to unseen categories.
In specific, we couple a query-based Transformer~\cite{vaswani2017attention} decoder, which generates instance mask proposals, with an image encoder, which is trained to output dense features (\ie patch tokens) aligned with text embeddings for a set of concepts from a \textit{frozen} CLIP~\cite{radford2021icml} text encoder.
Notably, the design of the proposed approach allows to do inference for both semantic and instance segmentations with strong performance compared to prior state-of-the-art approaches.

In summary, our contributions are three-fold:
(i) We introduce a challenging task, namely, zero-shot unsupervised transfer instance segmentation, which aims to segment object instances without human supervision or access to a target data distribution;
(ii) We propose a simple yet effective framework, termed \methodName, 
that goes beyond prior USSLIP approaches, 
and enables to concurrently perform instance segmentation in addition to semantic segmentation;
(iii) We show that \methodName performs favourably against state-of-the-art methods on standard unsupervised segmentation benchmarks (\eg, COCO~\cite{lin2014microsoft}, ImageNet-S~\cite{gao2021arxiv}) by a large margin in both zero-shot transfer and unsupervised domain adaptation settings.
\section{Related work}
\vspace{-2mm}
Our work relates to diverse themes in the literature including \textit{zero-shot semantic/instance segmentation}, \textit{unsupervised semantic segmentation} (with and without language-image pretraining), \textit{unsupervised object segmentation}, \textit{class-agnostic unsupervised instance segmentation}, \textit{universal architectures}, and \textit{open-vocabulary segmentation}. 

\mypara{Zero-shot semantic/instance segmentation with (image-)language pretraining.}
Zero-shot semantic/instance segmentation aims to generalise to unseen categories after training for seen categories with ground-truth annotations. 
The dominant approach exploits the relationships between category label embeddings produced by a language model (\eg, word2vec~\cite{mikolov2013efficient} or GloVe~\cite{pennington2014glove})~\cite{zhao2017open,bucher2019neurips,gu2020acmmm,xian2019semantic,kato2019zero,hu2020uncertainty,li2020consistent,pastore2021closer, zheng2021cvpr} to facilitate generalisation.
More recently, there has been growing interest in leveraging the joint image-text embedding space produced by a pretrained vision-language model (e.g. CLIP) to enable dense predictions~\cite{rao2021denseclip,li2022iclr,xu2021simple, ding2022cvpr,luddecke2022image}.
In a similar vein, we build our approach on a pretrained vision-language model to enable generalisation to novel categories, but with two key differences.
First, we do not assume access to a target data distribution, a setting termed \textit{zero-shot transfer} in~\cite{radford2021icml}.
Second, we do not use any manual annotations during training.
Note that the ``annotation free'' variant of MaskCLIP~\cite{zhou2022maskclip} enables semantic segmentation in a similar regime in which neither access to the target distribution nor ground-truth annotations are available.
We compare our method to MaskCLIP on semantic segmentation tasks in~\cref{sec:experiments}.

\mypara{Unsupervised semantic segmentation.}
A rich line of work has considered the problem of unsupervised semantic segmentation, creatively constructing learning signals from proxy tasks~\cite{ji2019iccv,ouali2020eccv,zhang2020self,van2021iccv,cho2021cvpr,hamilton2022iclr,vobecky2022arxiv}.
One practical challenge associated with these approaches is their reliance on a matching stage to enable deployment (typically performed with Hungarian matching~\cite{kuhn1955hungarian} on pixel-level segmentation annotations) that establishes correspondences between segments and category names.
By contrast, \methodName requires no access to pixel-level supervision during either training or inference.
Furthermore, unlike the above, \methodName is capable of instance-level predictions as well as semantic segmentation. 
We note one exception: MaskDistill~\cite{gansbeke2022maskdistill} also reports on unsupervised instance segmentation in addition to semantic segmentation (also using Hungarian matching to assign categories to predictions). 
In~\cref{sec:experiments}, we compare \methodName with MaskDistill on unsupervised instance segmentation.

\mypara{Unsupervised semantic segmentation with language-image pretraining (USSLIP).}
To achieve independence from pixel-level annotations during both training and inference, a recent line of work targeting unsupervised semantic segmentation proposes to leverage a vision-language model (\eg, CLIP~\cite{radford2021icml}) to assign names to categories~\cite{shin2022reco,shin2022namedmask}.
To do so, images are curated from an unlabelled image collection using the retrieval abilities of the vision-language model, and then segmented via co-segmentation~\cite{shin2022reco} or salient object detection~\cite{shin2022namedmask}.
However, while these methods avoid pixel-level annotations, they are either fragile (\ie, co-segmentation used in~\cite{shin2022reco}) or rigid (\ie, a new segmenter needs to be retrained from scratch for each new category~\cite{shin2022namedmask}).
Moreover, no USSLIP method to date supports instance segmentation.
\methodName builds on this line of work, but addresses its limited functionality by enabling instance segmentation, and improves both robustness and flexibility.

\mypara{Unsupervised object segmentation.}
Unsupervised object segmentation, also referred to as saliency detection, aims to train a detector to segment prominent object regions in images without human supervision.
Traditionally, handcrafted methods have been proposed utilising low-level cues such as centre prior~\cite{judd2009iccv}, contrast prior~\cite{itti1998tpami}, and boundary prior~\cite{wei2012eccv}.
A more recent line of research uses objectness properties emerging from self-supervised features extracted from modern vision architectures~\cite{simeoni2021localizing, wang2022cvpr, shin2022unsupervised} (\ie DINO~\cite{caron2021iccv}).
In this work, we adopt SelfMask~\cite{shin2022unsupervised} to generate object masks for images that are used as pseudo-masks for our training.

\mypara{Class-agnostic unsupervised instance segmentation.} Recently, FreeSOLO~\cite{wang2022freesolo} proposed a self-supervised framework for the class-agnostic instance segmentation task. For this, coarse object masks are first generated by using the object localisation property of self-supervised features (e.g., DenseCL~\cite{wang2020DenseCL}), then a class-agnostic object detector is trained with the initial masks via a self-training scheme~\cite{wang2022freesolo}. Concurrent work, CutLER~\cite{wang2023cut}, follows the similar framework, but with better initial masks produced by proposed MaskCut.
Unlike the above, we focus on the conventional \textit{class-aware} instance segmentation.
The classification of each instance mask is made possible as \methodName is built on the recent progress in unsupervised semantic segmentation with language-image pretraining (i.e., ReCo~\cite{shin2022reco}).

\mypara{Universal architectures.}
Recently, universal architectures that deliver multiple object detection/segmentation tasks in a unified manner have gained considerable attention~\cite{carion2020eccv,cheng2021per,zhang2021knet,cheng2021mask2former}.
Similarly, we propose an architecture that can tackle both semantic and instance segmentations with a single architecture with two key differences: (i) \methodName is flexible in terms of categories to segment as we use a text encoder as a classifier;
(ii) unlike the above which need to train a model from scratch for a different task, \methodName requires only a single training for semantic and instance segmentations. 

\mypara{Open-vocabulary segmentation.} Increasing the number of object categories to be segmented has been explored by utilising
class-incremental few-shot learning~\cite{hu2020learning},
captions~\cite{ghiasi2021open},
grounded text descriptions~\cite{kamath2021mdetr},
as well as
annotation transfer~\cite{hu2018learning,huynh2021open}
and
pairwise class balance regularisation~\cite{he2022relieving}. 
Similarly, we seek to scale the number of classes to be segmented, but without human supervision.

\begin{figure*}[!th]
    \centering
    \includegraphics[width=.95\textwidth]{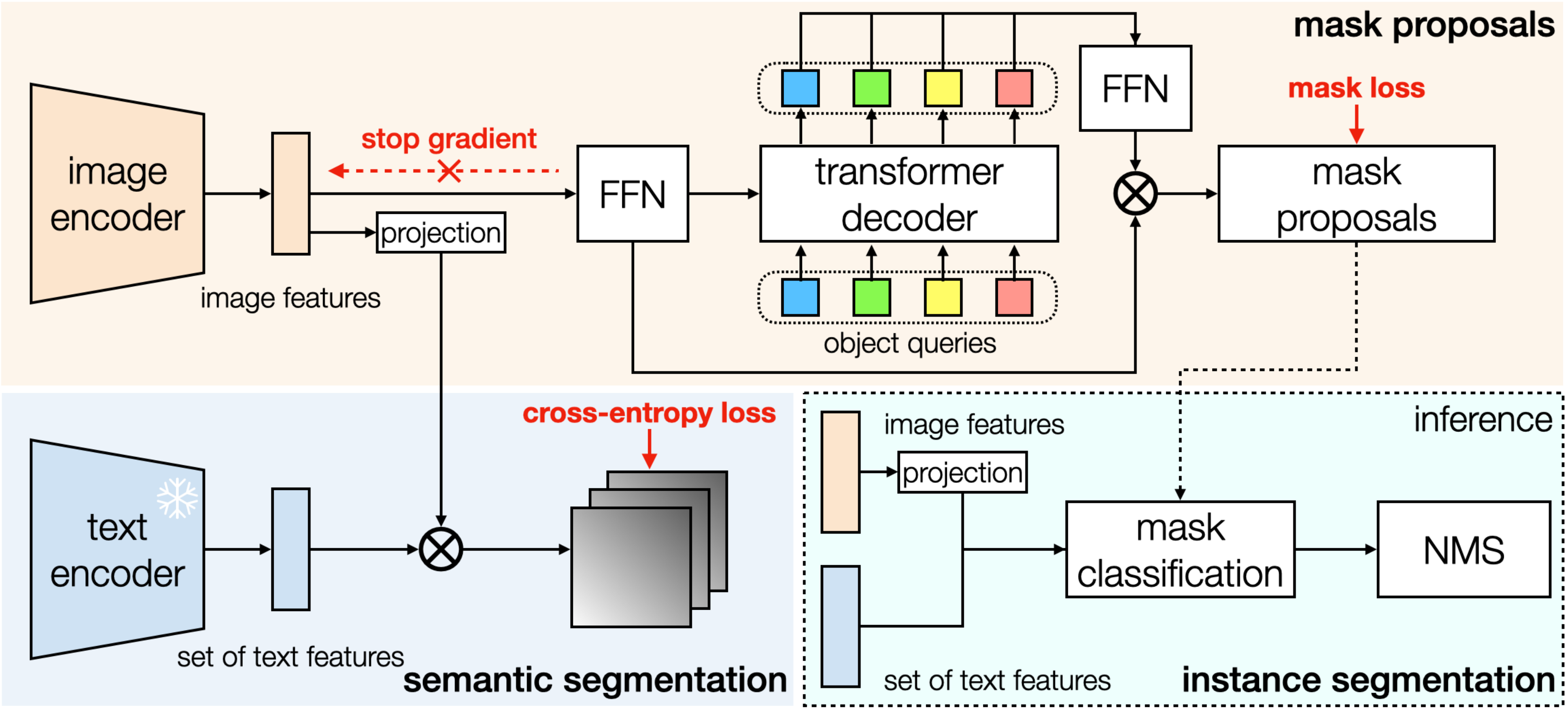}
    \vspace{-3mm}
    \caption{\textbf{Overview of \methodName}. Given an image encoder and a text encoder from a language-image model (\eg, CLIP), \methodName learns to perform both semantic and instance segmentation.
    (Top) image features for an image are fed to a feed-forward network (FFN) followed by a transformer decoder to produce mask proposals, which are used to make predictions for instance segmentation at inference (bottom right). 
    At the same time, the image features are projected into a text embedding space in which semantic predictions are made via a dot-product between the projected image features and frozen text features for a set of categories (bottom left).
    For simplicity, the pseudo-mask generation step is omitted. 
    See the text for details.}
    \vspace{-6mm}
    \label{fig:overview}
\end{figure*}

\section{Method}\label{sec:method}
\vspace{-2mm}
In this section,
we start by introducing the considered problem scenario, 
namely, zero-shot unsupervised transfer instance segmentation in~\cref{subsec:formulation},
and describe the core building blocks of our proposed approach in~\cref{subsec:pseudo-mask-generation}, followed by the architecture details for addressing unsupervised semantic and instance segmentation tasks with pretrained language-image models in~\cref{subsec:zesti}.

\subsection{Problem scenario}
\label{subsec:formulation}
\vspace{-2mm}
We consider the problem of zero-shot unsupervised transfer instance segmentation, which aims to jointly segment objects present in an image and predicts their semantic categories in both unsupervised and zero-shot transfer manner. 
The unsupervised property of the task prohibits any reliance on manual supervision for instance segmentation, while the zero-shot transfer property assumes that a segmenter has no access to a target data distribution (e.g., a training split of an evaluation benchmark). 
Note that, such properties pose significant differences from the existing zero-shot instance segmentation, which leverages human supervision (e.g., pixel-level annotations) in a training split (of an evaluation benchmark) for a certain group of classes (e.g., seen categories) during training.

To tackle this challenge, we propose a simple yet effective framework in which we first predict class-agnostic object masks (mask proposal), 
then classify each mask (mask classification) based on the pixelwise classification obtained in a joint image-text space.
Formally, we seek to train a segmenter $\Phi_{\text{seg}}$, consisting of an image encoder $\Phi_{\mcal{I}}^{\text{enc}}$, an image decoder $\Phi_{\mcal{I}}^{\text{dec}}$, 
and a text encoder $\Phi_\mcal{T}$. 
The segmenter ingests an image $x~\in \mbb{R}^{3 \times H \times W}$, and a set of concepts/object categories~($\mcal{C}$), 
and outputs a set of masks for semantic segmentation (SS) and instance segmentation (IS):
\begin{align}
\Phi_{\text{seg}}(x, \mcal{C})=\begin{cases}
    \Phi_\mcal{T}(\mcal{C}) \mbf{W} \Phi_{\mcal{I}}^{\text{enc}}(x) \in \{0, 1\}^{|\mcal{C}|\times H \times W} \text{for SS},\\
    \Phi_{\mcal{I}}^{\text{dec}} \circ \Phi_{\mcal{I}}^{\text{enc}}(x) \in \{0, 1\}^{n\times H \times W} \text{ for IS}.
\end{cases}
\end{align}
where $\mbf{W}$ is a matrix projecting image features into a text embedding space 
and $n$ denotes a pre-defined number of object mask predictions.
Note that, at this stage, the object masks from the image decoder are class-agnostic.
To decide a class of the mask proposals, each mask is assigned a category via a dot-product between its average image embedding (from the image encoder) and text embeddings followed by a softmax (detailed in \cref{subsec:zesti}). It is worth noting that the design of our framework allows the model to tackle both instance and semantic segmentations concurrently---we show performance of our model on both tasks in~\cref{sec:experiments}.

In the following sections, we introduce the key components for our framework in an unsupervised and zero-shot transfer fashion: generating pseudo-labels for unlabelled images with existing pretrained foundation models, and an efficient transformer-based architecture for simultaneous semantic and instance segmentation.

\subsection{Pseudo-label training}\label{subsec:pseudo-mask-generation}
\vspace{-2mm}
To train a segmenter without relying on manual labels, we adopt pseudo-label training as in~\cite{shin2022reco, shin2022namedmask}. Here, we briefly describe our pseudo-mask generation process, composed of archive construction, unsupervised saliency detection, and copy-paste augmentation used to generate synthetic images containing multiple objects.

\mypara{Archive construction.} Given an image encoder \imageEncoder and a text encoder \textEncoder from a pretrained vision-language model (\eg, CLIP), we first build archives of images for a set of categories $\mcal{C}$ by curating images for each concept from an unlabelled image dataset~\indexDataset (called an index dataset). Formally, we extract a set of normalised image embeddings $\mcal{F}_\mcal{I}$ as follows:
\begin{equation}
    \mcal{F}_\mcal{I} = \{\phi_\mcal{I}(x_i) \in \mbb{R}^{d},~i=1, ..., N\}
\end{equation}
where $x_i \in \mbb{R}^{3\times H \times W}$ and $N$ denotes the total number of images in \indexDataset. Similarly, we extract a set of normalised text embeddings $\phi_\mcal{T}(c) \in \mbb{R}^{d}$ for a name of each category $c \in \mcal{C}$ from the text encoder. Then we select $k$ images with highest similarities between image and text embeddings to form an archive for a category $c$:
\begin{equation}
    \mcal{U}_c = \{x_i \in \mcal{U}~|~i \in \text{argtop}_k[\mcal{F}_\mcal{I} \phi_\mcal{T}(c)]) \}
\end{equation}
where $\text{argtop}_k$ returns indices of $k$ largest values.

\mypara{Unsupervised saliency detection.} Given the image archives for the categories of interest, we generate category-agnostic saliency masks $\mcal{S}_i \in \{0, 1\}^{H \times W}$ by feeding each image $x_i$ into an unsupervised saliency detector (\eg, SelfMask~\cite{shin2022unsupervised}). 
We then assign the corresponding category name~(from archive construction) and an instance id to the inferred saliency mask, which allows for training a segmenter for semantic and instance segmentations as described in~\cref{subsec:zesti}.

\mypara{Synthetic image generation with copy-paste augmentation.} To train a segmenter which can segment multiple objects within an image, we follow~\cite{shin2022namedmask} and use copy-paste augmentation~\cite{ghiasi2021cvpr} to synthesise an image with multiple objects. A pseudo-mask is created accordingly by copy-pasting the binary pseudo-masks of the images used for the synthetic image, with a unique instance id and a category label allocated to each mask. 

\subsection{Architecture}\label{subsec:zesti}
\vspace{-2mm}
To tackle both semantic and instance segmentation tasks while preserving zero-shot ability of a pretrained vision-language model (VLM), we propose a simple framework termed, \methodName, which operates on features from image and text encoders of VLM (shown in~\cref{fig:overview}). 

\mypara{Semantic segmentation.} Given an image encoder $\psi_\mcal{I}$ and a text encoder $\psi_\mcal{T}$ from a pretrained VLM, we extract dense features $\psi_\mcal{I}(x_i) \in \mbb{R}^{e_v \times h \times w}$ (\eg, patch tokens from a vision transformer) for an image $x_i$ from the image encoder where $e_v, h,$ and $w$ denote the dimensionality of a visual embedding space, height and width of the features, respectively. The dense features are projected into a text embedding space by a projection matrix $\mbf{W} \in \mbb{R}^{e_t \times e_v}$, where $e_t$ is a dimension of the text space. With text embeddings $\psi_\mcal{T}(\mcal{C}) \in \mbb{R}^{|\mcal{C}|\times e_t}$ from the text encoder for a set of categories, we compute logits via dot-product between the projected image features and text features which are followed by a softmax function:
\begin{equation}
    P_i = \text{softmax}(\psi_\mcal{T}(\mcal{C}) \widetilde{\psi}_\mcal{I}(x_i),~\text{\qcr{dim=0}})
\end{equation}
where $\widetilde{\psi}_\mcal{I}(.)$ and $P_i$ denote $\mbf{W}\psi_\mcal{I}(.)$ and the probability map, respectively. 
Then a cross-entropy loss $\mcal{L}_{ce}$ is used to minimise differences between the prediction and the corresponding pseudo-mask generated in~\cref{subsec:pseudo-mask-generation}. To inherit the zero-shot ability of pretrained VLM, 
we only optimise the parameters of the image encoder, leaving the text encoder frozen.
Note that this approach is related to MaskCLIP, but with a key difference: 
we update the parameters in image encoder, while MaskCLIP keeps the parameters fixed and instead uses value features from the last self-attention layer of the image encoder to produce a semantic prediction. We compare our method to MaskCLIP in~\cref{subsec:main-results}.

\mypara{Instance segmentation.}
Here, we first produce object mask proposals using a query-based transformer decoder. In detail, given dense image features $\psi_\mcal{I}(x_i)$ before projection to the textual space, we pass the features to a feed-forward network (FFN) with a hidden layer (\eg, an MLP with three layers) whose output features are used as values $V \in \mbb{R}^{d \times h \times w}$ for the transformer decoder. 
Given $n_q$ object queries $Q \in \mbb{R}^{n_q \times d}$ and $V$, the decoder outputs query vectors that are fed into another FFN before producing mask proposals $\mcal{M} \in \mbb{R}^{n_q \times h \times w}$ via a dot-product between the resulting $Q$ and $V$.
Then, we update the model with a bipartite matching loss~\cite{carion2020eccv, cheng2021per, cheng2021mask2former} $\mcal{L}_{mask}$ between the proposals and the pseudo-masks for the image. 
We find that it is essential to {\bf stop gradients} from the transformer decoder flowing to the image encoder, otherwise the model fails to converge (see~\cref{subsec:ablation-study}).
For $\mcal{L}_{mask}$, we use a mixture of dice coefficient~\cite{miletari2016vnet} and binary cross-entropy losses \(\mcal{L}_{mask} = \mcal{L}_{dice} + \mcal{L}_{bce}\) with equal weights following~\cite{cheng2021mask2former}.

During inference, we assign each mask proposal $m_l \in \mcal{M}$ a category whose text embedding shares the highest similarity with an average image embedding of the mask. For this, we first binarise $m_l$ with a threshold $t$ and compute the average image embedding $\overline{\psi}_\mcal{I}(x_i, m_l; t)$:
\begin{equation}
    \overline{\psi}_\mcal{I}(x_i, m_l; t) = \text{mean}(\widetilde{\psi}_\mcal{I}(x_i)[m_l > t]) \in \mbb{R}^{e_t}.
\end{equation}
Then, we assign the mask that category with highest similarity to the average image embedding:
\begin{equation}
    \underset{c \in \mcal{C}}{\text{argmax}}\big[\psi_\mcal{T}(\mcal{C}) \overline{\psi}_\mcal{I}(x_i, m_l; t)\big].
\end{equation}
Note that both text and average image embeddings are L2-normalised before dot-product. 
In addition, we compute a confidence score $s_l \in [0, 1]$ for each mask proposal, defined as the average value of the mask region multiplied by the maximum class probability for the mask (similarly to~\cite{cheng2021mask2former}). 
Lastly, to reduce false positives occurring from redundant predictions for a single object, we apply non-maximum suppression (NMS) to the proposals before outputting final instance predictions. 
We show the effect of NMS in~\cref{subsec:ablation-study}.

\vspace{2pt}
\noindent \textbf{Discussion.} 
The key differences of \methodName from prior work for unsupervised semantic segmentation with language-image pretraining are two-fold:
(i) rather than using a fixed n-way classifier, we use a pretrained, 
frozen text encoder as a classifier, 
and optimise an image encoder to output dense features aligned with the textual features from the text encoder,
a design choice that allows the model to be open-vocabulary;
(ii) we enable instance segmentation by training a query-based transformer decoder by bootstrapping the results from saliency detection via copy-paste augmentation.

\section{Experiments}\label{sec:experiments}
\vspace{-2mm}
In this section, we first describe the details of our experiments including datasets, 
network architecture, training and inference settings, and evaluation metrics. 
Next, we ablate components of our method such as the use of stop-gradient and non-maximum suppression, and report the performance of the model on both semantic and instance segmentation.

\subsection{Implementation details}
\vspace{-2mm}
\mypara{Datasets.} 
We evaluate our model on
COCO2017~\cite{lin2014microsoft} \qcr{val} split,
PASCAL VOC2012~\cite{Everingham10}, 
CoCA~\cite{zhang2020gicd}, and ImageNet-S~\cite{gao2021arxiv} \qcr{test} split for semantic segmentation and COCO-20K~\cite{vo2022eccv} for instance segmentation following~\cite{gansbeke2022maskdistill}.
To demonstrate our model's zero-shot ability to \textbf{new concepts}, 
we additionally consider the CUB-200-2011~\cite{WahCUB_200_2011} \qcr{test} split.
In detail, 
VOC2012 \qcr{trainval} split has 2,913 images with 21 categories including a background. 
COCO2017 \qcr{val} and CoCA are composed of 5,000 and 1,295 images with 80 object categories and a background class. ImageNet-S \qcr{test} consists of 27,423 images with 919 object classes which are a subset of ImageNet1K~\cite{russakovsky2015imagenet} classes.
COCO-20K comprises 19,817 images from the COCO2014 \qcr{train} split with the same 80 object classes as COCO2017.
CUB-200-2011 \qcr{test} is composed of 5,794 images with 200 fine-grained bird breeds.

Note that, in this paper, we primarily consider the zero-shot transfer setting, in which the model has no access to training data sharing a data distribution with a downstream benchmark.
Thus, throughout our experiments, 
we use images retrieved from ImageNet1K (1.2M images) and PASS~\cite{asano21pass} (1.4M images) by the ViT-L/14@336px CLIP model to form an index image dataset except an experiment in the unsupervised domain adaptation setting for the ImageNet-S benchmark where we only retrieve ImageNet1K images. 
For prompt engineering, we average text embeddings from
85 templates to obtain a textual feature for a category  following~\cite{zhou2022maskclip, shin2022reco, shin2022namedmask}.
In all cases, we fix the number of images for an archive as 500 (\ie 500 images for a category) as in~\cite{shin2022namedmask}.

\mypara{Architecture.} We use transformer-based CLIP models for the image encoder (\eg, ViT-B/16) and text encoder.
We use 6 transformer layers for the transformer decoder and three-layer MLP for the FFN.
We feed patch tokens from the last layer of the image encoder to the decoder after bilinearly upsampling them by a factor of 2 to enable predictions at a higher resolution.

\mypara{Training.}
We compute the final loss $\mcal{L}$ for the model as \(\mcal{L} = \mcal{L}_{ce} + \lambda_{mask}\mcal{L}_{mask}\) with $\lambda_{mask}$ set to 1.0.
We optimise our model with the AdamW optimiser~\cite{loshchilov2018decoupled}, with an initial  learning rate of 5e-5 and a weight decay of 0.05. 
For the image encoder, we use a smaller learning rate of 5e-6. 
We train for 20K iterations with the Poly learning rate scheduler~\cite{chen2018eccv}, except when training for 919 categories of ImageNet-S where the model is updated for 80K iterations. 
We use standard data augmentations such as random resizing, cropping and colour jittering.
Following~\cite{shin2022namedmask}, we adopt copy-paste augmentation~\cite{ghiasi2021cvpr} and set the maximum number of images used for copy-pasting to 10.
To further encourage the model to differentiate objects of the same category, we select images used for copy-paste from a randomly selected archive 50\% of the time.
As in~\cite{carion2020eccv, cheng2021mask2former}, we compute a mask loss for predictions by each transformer decoder layer. 

\mypara{Inference.} 
We perform inference on images at their original resolution, except for the large-scale ImageNet-S benchmark where we resize images with a longer side larger than 1024 while preserving its aspect ratio. 
For such cases, the original resolution is restored with a bilinear upsampler following~\cite{shin2022namedmask}.
In addition, we apply NMS for instance segmentation predictions as mentioned in~\cref{subsec:zesti}.

\mypara{Evaluation metrics.}
To measure our model's performance, 
we use the standard metrics such as mean intersection-over-union (mIoU) for semantic segmentation and COCO-style mask average precision ($\maskap$) for instance segmentation.

\begin{figure}[!t]
    \centering
    \includegraphics[width=.48\textwidth]{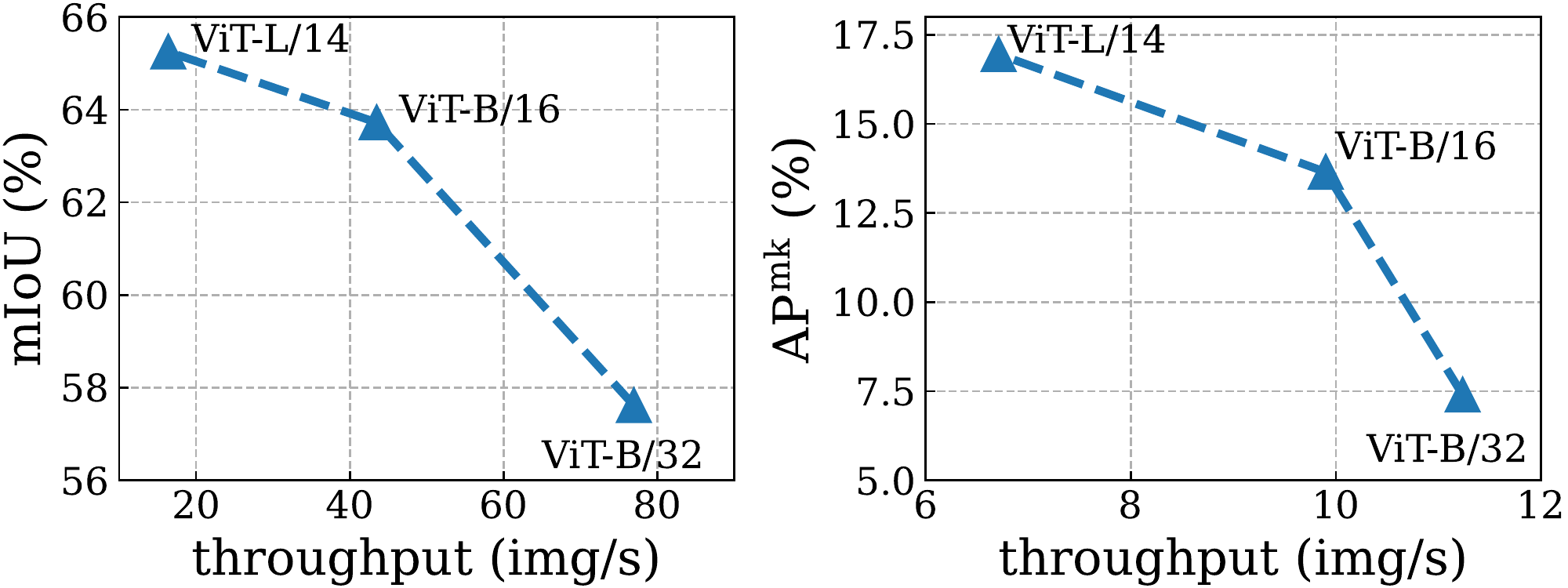}
    \vspace{-7mm}
    \caption{With more computation measured in throughput, 
    \methodName can produce better performance in both semantic segmentation (left) and instance segmentation (right).}
    \vspace{-5mm}
    \label{fig:scale-experiment}
\end{figure}

\subsection{Ablation study}
\label{subsec:ablation-study}

Here, we study the influence of the components in our method, including the choice of encoder architecture, stop-gradient and NMS. 
For experiments in the ablation study, we report the results on the VOC2012 \qcr{trainval} split.

\mypara{Effect of encoder architecture.} 
In~\cref{fig:scale-experiment}, we show the performance of our model with different transformer-based CLIP image encoders such as ViT-B/32, ViT-B/16, and ViT-L/14~\cite{dosovitskiy2020image}.\footnote{While there are also ResNet-based CLIP encoders, we found them not suitable for instance segmentation as they directly output features in the joint image-text space via an attention pooling~\cite{radford2021icml}.} 
We observe that at the cost of computation measured in throughput, heavier models consistently outperform lighter models in both mIoU (left) and $\maskap$ (right). 
While ViT-L/14 performs best, we report results with either ViT-B/32 or ViT-B/16 in the following experiments to limit differences in performance due to model size (ResNet50~\cite{he2016cvpr} is typically used by previous unsupervised methods). 
Note that ViT-B/32 and ViT-B/16 are the lightest CLIP models compatible with our framework.

\begin{table}[!t]
\centering
\begin{tabular}{c c | ccc}
stop-grad & NMS & $\maskap$ & $\maskapfifty$ & $\maskapseventyfive$\\ \thickhline
\xmark & \xmark & 0.3 & 0.4 & 0.3\\
\checkmark & \xmark & 4.4 & 8.7 & 4.2 \\
\checkmark & \checkmark & \textbf{13.7} & \textbf{30.9} & \textbf{11.1} \\
\end{tabular}
\vspace{-3mm}
\caption{Stopping gradients from the transformer decoder to the image encoder plays a crucial role in our framework while non-maximum suppression (NMS) produces a substantial boost in performance.
The performance is measured in COCO-style AP metrics for instance segmentation.
}
\label{tab:component-analysis}
\vspace{-3.5mm}
\end{table}
\setlength\tabcolsep{3.5pt}
\begin{table}[!t]
\centering
\begin{tabular}{c|ccc|ccc}
      & \multicolumn{3}{c|}{\textit{class-agnostic}} & \multicolumn{3}{c}{\textit{class-aware}} \\
stop-grad & $\maskap$ & $\maskapfifty$ & $\maskapseventyfive$ & $\maskap$ & $\maskapfifty$ & $\maskapseventyfive$\\ \thickhline
\xmark & 8.4 & 19.9 & 6.4 & 1.0 & 1.9 & 0.9 \\
\checkmark & 9.9 & 24.1 & 7.4 & 13.7 & 30.9 & 11.1  
\end{tabular}
\vspace{-3.5mm}
\caption{Applying a stop-grad operation between the image encoder and decoder allows the encoder features to keep semantic representations.}
\label{tab:stop-grad}
\vspace{-7mm}
\end{table}

\mypara{Effect of stop-gradient and non-maximum suppression.}
As described in~\cref{subsec:zesti}, we prevent gradients from backpropagating to the encoder parameters when optimising the transformer decoder to generate mask proposals. 
We observe in~\cref{tab:component-analysis} that this is crucial, otherwise the optimisation does not converge to a reasonable solution. 
For our model trained with stop-gradient, applying NMS to its mask proposals brings a noticeable gain in performance from 4.4 to 13.7 $\maskap$.
This is because the model tends to predict redundant mask proposals for a single object, increasing false positives during evaluation.
We therefore employ both stop-gradient and NMS throughout the experiments.

\mypara{Analysis on a stop-gradient operation.}
To further investigate why applying a stop-grad operation is essential in our framework, 
we hypothesise that if stop-grad is not applied, gradients from the mask loss for instance segmentation could potentially dominate the visual feature learning,
thus harming the visual-language alignment in pretrained VLM.
To verify this, we evaluate the models trained w/ and w/o the stop-grad w.r.t. class-agnostic and -aware $\maskap$. As shown in~\cref{tab:stop-grad}, while the model trained w/o stop-grad performs poorly on class-aware $\maskap$, it performs reasonably on class-agnostic $\maskap$, indicating that the resulting features are capable of segmenting objects but not suitable for classification. Thus, it is crucial to separate features used for semantic and instance predictions via a stop-grad for enabling class-aware instance segmentation.

\subsection{Main results}\label{subsec:main-results}
\vspace{-2mm}
Here, we compare \methodName to existing unsupervised instance segmentation and semantic segmentation approaches. While we mainly focus on the zero-shot transfer setting in which the model has no access to training data for the target downstream task, 
we also report results for semantic segmentation in the unsupervised domain adaptation setting to draw comparison with existing approaches, where the target data distribution is exposed to the model.

\mypara{Unsupervised instance segmentation.}
In~\cref{tab:instance-segmentation}, we evaluate our model on unsupervised instance segmentation on the COCO-20K dataset. 
As a baseline for our method, we evaluate MaskCLIP for instance segmentation by treating its semantic segmentation masks for an image as mask proposals. When comparing to the state-of-the-art approach~\cite{gansbeke2022maskdistill}, our model shows comparable (with ViT-B/32) or better performance (with ViT-B/16) by 3.3 $\maskapfifty$. For qualitative visualisations of our model's predictions, see~\cref{fig:teaser}.

\setlength\tabcolsep{4.5pt}
\begin{table}[!t]
\small
\centering
\begin{tabular}{l c ccc}
\toprule
model & backbone & $\maskap$ & $\maskapfifty$ & $\maskapseventyfive$ \\ \midrule
\multicolumn{5}{l}{\textit{unsupervised methods w/o language-image pretraining}} \\
DINO~\cite{caron2021iccv} & ViT-S/16 & 0.7        & 2.0         & 0.4        \\
LOST~\cite{simeoni2021localizing} & ViT-S/16 & 1.2        & 3.3         & 0.6        \\
MaskDistill~\cite{gansbeke2022maskdistill} & ViT-S/16 & 1.7        & 4.1         & 1.4        \\
MaskDistill~\cite{gansbeke2022maskdistill}$^\dagger$ & RN50-C4 & 3.5     & 7.7         & 2.9        \\ \midrule
\multicolumn{5}{l}{\textit{unsupervised method w/ language-image pretraining}}  \\
MaskCLIP~\cite{zhou2022maskclip} & ViT-B/32 & 0.3 & 0.8 & 0.2\\
\methodName (Ours)& ViT-B/32 & 3.4 & 8.0 & 2.6 \\ \hdashline
MaskCLIP~\cite{zhou2022maskclip} & ViT-B/16 & 1.3 & 3.4 & 0.8\\
\methodName (Ours)& ViT-B/16 & \textbf{5.7} & \textbf{11.0} & \textbf{5.4} \\ \bottomrule
\end{tabular}
\vspace{-3mm}
\caption{Comparison to previous unsupervised instance segmentation methods on COCO-20K.
$^\dagger$Mask R-CNN~\cite{he2017iccv} trained with pseudo-masks from MaskDistill. 
The numbers for the methods without language-image pretraining are quoted from~\cite{gansbeke2022maskdistill}.}
\vspace{-2mm}
\label{tab:instance-segmentation}
\end{table}

\setlength\tabcolsep{2pt}
\begin{table}[!t]
\small
\centering
\begin{tabular}{llcc}
\toprule
\multicolumn{1}{c}{model} &
  \multicolumn{1}{c}{arch.} &
  \multicolumn{1}{c}{COCO} &
  \multicolumn{1}{c}{CoCA} \\ \hline
\multicolumn{4}{l}{\textit{initialised with different encoder features}}\\
ReCo$^\dagger$~\cite{shin2022reco} & DeiT-S/16 \& RN50x16 & 23.8 & 28.8 \\
NamedMask$^\ddagger$~\cite{shin2022namedmask} & RN50 \& DLv3+ & 28.4 & 27.3 \\ \hline
\multicolumn{4}{l}{\textit{initialised with CLIP encoder features}}\\
MaskCLIP~\cite{zhou2022maskclip} & ViT-B/16 & 20.6 & 20.2 \\
\methodName (Ours) & ViT-B/16 & \textbf{32.8} & \textbf{32.7} \\ \bottomrule
\end{tabular}
\vspace{-3mm}
\caption{
Comparison to previous unsupervised semantic segmentation methods leveraging image-language pretraining on COCO and CoCA in terms of mIoU (\%). $^\dagger$Initialised with supervised Stylised-ImageNet pretraining~\cite{naseer2021intriguing}. $^\ddagger$Initialised with DINO~\cite{caron2021iccv}.
}
\vspace{-5mm}
\label{tab:semantic-segmentation}
\end{table}

\setlength\tabcolsep{3pt}
\begin{table*}[th]
\centering
\begin{tabular}{l c c c ccccc}
\toprule
model & arch. & \# params (M) & throughput (img/s) & mIoU & S & MS & ML & L\\ \midrule
\multicolumn{9}{l}{\textit{unsupervised methods w/o language-image pretraining}}\\
PASS$_p$~\cite{gao2021arxiv} & RN50 & 25.6 & - & 6.6 & 1.3 & 4.6 & 7.1 & 8.4\\
PASS$_s$~\cite{gao2021arxiv} & RN50 & 25.6 & - & 11.0 & 2.4 & 8.3 & 11.9 & 13.4\\ \midrule
\multicolumn{9}{l}{\textit{unsupervised methods w/ language-image pretraining}}\\
ReCo$^\dagger$~\cite{shin2022reco} & DeiT-S/16 \& RN50x16 & 170.4 & 32.3 & 10.3 & 6.0 & 11.6 & 10.2 & 6.7\\ 
NamedMask$^\ddagger$~\cite{shin2022namedmask} & RN50 \& DLv3+ & 26.6 & \textbf{125.0} & 22.9 & 5.1 & 19.4 & 24.4 & 19.8 \\
\methodName (Ours) & ViT-B/32 & 87.8 & 76.9 & 27.5 & 5.6 & 22.3 & 28.9 & 26.5\\
\methodName (Ours) & ViT-B/16 & 86.2 & 43.5 & \textbf{37.4} & \textbf{10.7} & \textbf{32.1} & \textbf{40.2} & \textbf{33.4} \\ \bottomrule
\end{tabular}
\vspace{-3mm}
\caption{
Comparison to existing unsupervised methods on the ImageNet-S benchmark with 919 object categories in the unsupervised domain adaptation setting.
We also show mIoU in diverse object sizes from small (S), medium-small (MS), medium-large (ML), and large (L).
$^\dagger$Encoder initialised with supervised Stylised-ImageNet pretraining. $^\ddagger$Encoder initialised with unsupervised pretraining (\ie, DINO).
}
\vspace{-5mm}
\label{tab:imagenet-s}
\end{table*}

\mypara{Unsupervised semantic segmentation.}
For semantic segmentation, we primarily compare to unsupervised approaches that leverage language-image pretraining such as NamedMask, ReCo, and MaskCLIP. 
As ReCo and MaskCLIP do not predict a ``background'' category, we re-implement their methods to predict background class since it appears in all the benchmarks considered in our experiments. Specifically, for ReCo, we follow~\cite{shin2022namedmask} and treat the pixels as background if their maximum class probability is lower than a threshold (=0.9). 
For MaskCLIP, we simply provide a text embedding for ``background'' along with other object category embeddings which we find more effective than thresholding.

In~\cref{tab:semantic-segmentation}, we evaluate our model in the zero-shot transfer setting on the COCO \qcr{val} and CoCA benchmarks and compare to unsupervised methods. 
Note that ReCo and NamedMask have different settings from ours in terms of initialisation and architecture for a backbone (\ie, ReCo initialises its backbone with supervised Stylised-ImageNet training and NamedMask with DINO), thus a direct comparison is not possible. 
Relative to MaskCLIP (which is comparable), our model shows improvements of 12.2 and 12.5 mIoU on COCO and CoCA, respectively.

In~\cref{tab:imagenet-s}, we evaluate our method in the unsupervised domain adaptation setting, where  the model is trained with images retrieved from the ImageNet1K \qcr{train} split, and evaluated on ImageNet-S. 
Compared to the state-of-the-art unsupervised method (\ie NamedMask), our approach achieves a gain of 4.6 mIoU with ViT-B/32 and 14.5 mIoU with ViT-B/16 at the expense of lower throughput.

\begin{table}[!t]
\centering
\begin{tabular}{cc}
\toprule
category-specific label & CUB-200-2011\\ \midrule
\xmark  & 72.5\\
\checkmark & 72.6\\ \bottomrule
\end{tabular}
\vspace{-6pt}
\caption{High-level to low-level zero-shot transfer on the CUB-200-2011 benchmark. When given a fine-grained bird breed, \methodName can segment the corresponding bird regions as good as when it is given a high-level category ``bird.''}
\vspace{-3.5mm}
\label{tab:high-to-low}
\end{table}

\mypara{Generalisation to new categories.}
Since we optimise our image encoder to produce visual embeddings aligned to the text embedding from the frozen text encoder,
we expect the resulting model to be capable of segmenting objects of novel concepts which are unseen during training.
To verify this, we consider two scenarios: 
(i) a high-level to low-level category transfer, 
{\em i.e.}, the model is evaluated on categories that it did not encounter during training but only their superset category;
(ii) transfer to unseen categories semantically far from those it has seen during training.

For the first scenario, we evaluate our model, trained for 80 categories in COCO including `bird', on the \qcr{test} split of CUB-200-2011 benchmark which has 200 fine-grained bird categories.
Here, given a high-level category (\ie ``bird'') or a low-level category for an image (\ie image-specific fine-grained bird categories), we encode the category with the text encoder and proceed with segmentation as usual.
Then we compare the segmentation result with the groundtruth mask.
It is worth mentioning that the performance is measured in IoU rather than mIoU, as we do not expect the model to distinguish between the fine-grained categories. 
This is because the image archive for ``bird'' is likely to contain images of different bird breeds, which encourages the model to learn the invariance between birds. 
However, we expect the model to also identify the ``bird'' regions given a specific bird breed as a target. %
As shown in~\cref{tab:high-to-low}, when given low-level categories (bird breeds) as input, 
our model can perform equally well as when given a high-level category (``bird'').
Note that the semantically closest category to the fine-grained categories among the 80 classes in COCO is ``bird'' and that only 16 out of 200 fine-grained bird categories contain ``bird'' as a part of its name (\eg ``Anna Hummingbird''). This implies that the model is equipped with knowledge about fine-grained bird species.

For the second scenario, we split 65 seen and 15 unseen classes in the COCO dataset and evaluate our model on the unseen classes following prior work on zero-shot instance segmentation~\cite{zheng2021cvpr}. For this, we train our model with image archives constructed only for the seen categories.
As shown in~\cref{tab:unseen-instance-segmentation}, when compared to a baseline unsupervised method, our model performs favourably by a notable margin (i.e., $+$5.2 $\maskapfifty$). It indicates good generalisability of our model to the unseen categories which are semantically remote from the concepts seen during training.

\setlength\tabcolsep{6pt}
\begin{table}[!t]
\centering
\begin{tabular}{l|lll}
\multicolumn{1}{c|}{model} & \multicolumn{1}{c}{$\maskap$} & \multicolumn{1}{c}{$\maskapfifty$} & \multicolumn{1}{c}{$\maskapseventyfive$} \\ \thickhline
MaskCLIP~\cite{zhou2022maskclip} & 0.7 & 2.0  & 0.4  \\
\methodName (Ours) & 3.3 \green{($+2.6$)} & 7.2 \green{($+5.2$)} & 2.8 \green{($+2.4$)}
\end{tabular}
\vspace{-6pt}
\caption{Zero-shot unsupervised instance segmentation for 15 unseen categories on COCO-20K.}
\vspace{-6mm}
\label{tab:unseen-instance-segmentation}
\end{table}

\subsection{Limitations}
\vspace{-2mm}
Although we show strong performance on unsupervised semantic segmentation and instance segmentation leveraging only image-language pretraining, we also note two limitations to our work:
(i) While the use of a vision-language model such as CLIP~\cite{radford2021icml} significantly simplifies deployment relative to unsupervised approaches that employ Hungarian matching, it also represents the source of potential error.
For instance, our method will be unable to segment categories that are not present in CLIP's pretraining data (extremely rare concepts, for example).
(ii) Our pipeline for sourcing pseudo-masks with a vision-language model (VLM) and a saliency detector has drawbacks. Images retrieved for a given concept by VLM can contain objects of a distracting class which the detector can highlight together with the desired category object. 
For example, both a skateboard and a person riding it can be segmented by the detector when we retrieve the image for ``skateboard'' and generate a pseudo-mask for it. 
\vspace{-2mm}
\section{Broader impact}
\vspace{-2mm}
The goal of this work is to propose a practical framework for instance and semantic segmentation.
As such, we hope that our work facilitates many useful applications of segmentation (medical image analysis, fault detection in manufacturing, security monitoring etc.).
However, automatic segmentation represents a dual-use technology and is therefore subject to misuse (unlawful surveillance, for example).
We also note that we build \methodName on top of foundation models like CLIP~\cite{radford2021icml}.
These models are known to reflect biases present in large, minimally curated internet corpora and thus our model is likely to inherit these biases also.
Consequently, any practical deployment of \methodName will require assessment (and potentially also mitigation) of the risks posed by such biases.
\vspace{-2mm}
\section{Conclusion}
\vspace{-2mm}
In this work, we introduced \methodName, 
the first framework for joint instance segmentation and semantic segmentation in a zero-shot transfer setting that requires no pixel-level or instance-level annotation.
We employ a query-based transformer architecture for instance segmentation and train it on pseudo-labels generated from applying an unsupervised saliency detector to images retrieved by CLIP. 
Through careful experiments, we demonstrated the effectiveness of \methodName across both instance segmentation and semantic segmentation tasks.
In future work, we intend to explore the application of \methodName to other modalities such as video.

\mypara{Acknowledgements and disclosure of funding.}
This work was performed using resources provided by the Cambridge Service for Data Driven Discovery (CSD3) operated by the University of Cambridge Research Computing Service (www.csd3.cam.ac.uk), provided by Dell EMC and Intel using Tier-2 funding from the Engineering and Physical Sciences Research Council (capital grant EP/T022159/1), and DiRAC funding from the Science and Technology Facilities Council (www.dirac.ac.uk).
GS is supported by AI Factory, Inc. in Korea.
GS would like to thank Guanqi Zhan for proof-reading and Zheng Fang for the enormous support.
SA would like to acknowledge the support of Z. Novak and N. Novak in enabling his contribution.

{\small
\bibliographystyle{ieee_fullname}
\bibliography{refs}

\begin{thebibliography}{10}\itemsep=-1pt

\bibitem{asano21pass}
Yuki~M. Asano, Christian Rupprecht, Andrew Zisserman, and Andrea Vedaldi.
\newblock Pass: An imagenet replacement for self-supervised pretraining without
  humans.
\newblock {\em NeurIPS Track on Datasets and Benchmarks}, 2021.

\bibitem{bilic2019liver}
Patrick Bilic, Patrick~Ferdinand Christ, Eugene Vorontsov, Grzegorz Chlebus,
  Hao Chen, Qi Dou, Chi-Wing Fu, Xiao Han, Pheng-Ann Heng, J{\"u}rgen Hesser,
  et~al.
\newblock The liver tumor segmentation benchmark (lits).
\newblock {\em arXiv preprint arXiv:1901.04056}, 2019.

\bibitem{bommasani2021opportunities}
Rishi Bommasani, Drew~A Hudson, Ehsan Adeli, Russ Altman, Simran Arora, Sydney
  von Arx, Michael~S Bernstein, Jeannette Bohg, Antoine Bosselut, Emma
  Brunskill, et~al.
\newblock On the opportunities and risks of foundation models.
\newblock {\em arXiv:2108.07258}, 2021.

\bibitem{bucher2019neurips}
Maxime Bucher, Tuan-Hung Vu, Mathieu Cord, and Patrick P{\'e}rez.
\newblock Zero-shot semantic segmentation.
\newblock In {\em NeurIPS}, 2019.

\bibitem{carion2020eccv}
Nicolas Carion, Francisco Massa, Gabriel Synnaeve, Nicolas Usunier, Alexander
  Kirillov, and Sergey Zagoruyko.
\newblock End-to-end object detection with transformers.
\newblock In {\em ECCV}, 2020.

\bibitem{caron2021iccv}
Mathilde Caron, Hugo Touvron, Ishan Misra, Herv\'e J\'egou, Julien Mairal,
  Piotr Bojanowski, and Armand Joulin.
\newblock Emerging properties in self-supervised vision transformers.
\newblock In {\em ICCV}, 2021.

\bibitem{chen2018eccv}
Liang-Chieh Chen, Yukun Zhu, George Papandreou, Florian Schroff, and Hartwig
  Adam.
\newblock Encoder-decoder with atrous separable convolution for semantic image
  segmentation.
\newblock In {\em ECCV}, 2018.

\bibitem{cheng2021mask2former}
Bowen Cheng, Ishan Misra, Alexander~G. Schwing, Alexander Kirillov, and Rohit
  Girdhar.
\newblock Masked-attention mask transformer for universal image segmentation.
\newblock In {\em CVPR}, 2022.

\bibitem{cheng2021per}
Bowen Cheng, Alex Schwing, and Alexander Kirillov.
\newblock Per-pixel classification is not all you need for semantic
  segmentation.
\newblock {\em NeurIPS}, 2021.

\bibitem{cho2021cvpr}
Jang~Hyun Cho, Utkarsh Mall, Kavita Bala, and Bharath Hariharan.
\newblock Picie: Unsupervised semantic segmentation using invariance and
  equivariance in clustering.
\newblock In {\em CVPR}, 2021.

\bibitem{cordts2016cvpr}
Marius Cordts, Mohamed Omran, Sebastian Ramos, Timo Rehfeld, Markus Enzweiler,
  Rodrigo Benenson, Uwe Franke, Stefan Roth, and Bernt Schiele.
\newblock The cityscapes dataset for semantic urban scene understanding.
\newblock In {\em CVPR}, 2016.

\bibitem{ding2022cvpr}
Jian Ding, Nan Xue, Gui-Song Xia, and Dengxin Dai.
\newblock Decoupling zero-shot semantic segmentation.
\newblock In {\em CVPR}, 2022.

\bibitem{dosovitskiy2020image}
Alexey Dosovitskiy, Lucas Beyer, Alexander Kolesnikov, Dirk Weissenborn,
  Xiaohua Zhai, Thomas Unterthiner, Mostafa Dehghani, Matthias Minderer, Georg
  Heigold, Sylvain Gelly, Jakob Uszkoreit, and Neil Houlsby.
\newblock An image is worth 16x16 words: Transformers for image recognition at
  scale.
\newblock In {\em ICLR}, 2021.

\bibitem{Everingham10}
M. Everingham, L. Van~Gool, C.~K.~I. Williams, J. Winn, and A. Zisserman.
\newblock The pascal visual object classes (voc) challenge.
\newblock {\em IJCV}, 2010.

\bibitem{gao2021arxiv}
Shanghua Gao, Zhong-Yu Li, Ming-Hsuan Yang, Ming-Ming Cheng, Junwei Han, and
  Philip Torr.
\newblock Large-scale unsupervised semantic segmentation.
\newblock {\em arXiv:2106.03149}, 2021.

\bibitem{geirhos2018iclr}
Robert Geirhos, Patricia Rubisch, Claudio Michaelis, Matthias Bethge, Felix~A.
  Wichmann, and Wieland Brendel.
\newblock Imagenet-trained {CNN}s are biased towards texture; increasing shape
  bias improves accuracy and robustness.
\newblock In {\em ICLR}, 2019.

\bibitem{ghiasi2021cvpr}
Golnaz Ghiasi, Yin Cui, Aravind Srinivas, Rui Qian, Tsung-Yi Lin, Ekin~D.
  Cubuk, Quoc~V. Le, and Barret Zoph.
\newblock Simple copy-paste is a strong data augmentation method for instance
  segmentation.
\newblock In {\em CVPR}, 2021.

\bibitem{ghiasi2021open}
Golnaz Ghiasi, Xiuye Gu, Yin Cui, and Tsung-Yi Lin.
\newblock Open-vocabulary image segmentation.
\newblock {\em arXiv:2112.12143}, 2021.

\bibitem{gu2020acmmm}
Zhangxuan Gu, Siyuan Zhou, Li Niu, Zihan Zhao, and Liqing Zhang.
\newblock Context-aware feature generation for zero-shot semantic segmentation.
\newblock In {\em ACM MM}, 2020.

\bibitem{hamilton2022iclr}
Mark Hamilton, Zhoutong Zhang, Bharath Hariharan, Noah Snavely, and William~T.
  Freeman.
\newblock Unsupervised semantic segmentation by distilling feature
  correspondences.
\newblock In {\em ICLR}, 2022.

\bibitem{he2017iccv}
Kaiming He, Georgia Gkioxari, Piotr Dollár, and Ross Girshick.
\newblock Mask r-cnn.
\newblock In {\em ICCV}, 2017.

\bibitem{he2016cvpr}
Kaiming He, Xiangyu Zhang, Shaoqing Ren, and Jian Sun.
\newblock Deep residual learning for image recognition.
\newblock In {\em CVPR}, 2016.

\bibitem{he2022relieving}
Yin-Yin He, Peizhen Zhang, Xiu-Shen Wei, Xiangyu Zhang, and Jian Sun.
\newblock Relieving long-tailed instance segmentation via pairwise class
  balance.
\newblock {\em arXiv:2201.02784}, 2022.

\bibitem{hu2020uncertainty}
Ping Hu, Stan Sclaroff, and Kate Saenko.
\newblock Uncertainty-aware learning for zero-shot semantic segmentation.
\newblock In {\em NeurIPS}, 2020.

\bibitem{hu2018learning}
Ronghang Hu, Piotr Doll{\'a}r, Kaiming He, Trevor Darrell, and Ross Girshick.
\newblock Learning to segment every thing.
\newblock In {\em CVPR}, 2018.

\bibitem{hu2020learning}
Xinting Hu, Yi Jiang, Kaihua Tang, Jingyuan Chen, Chunyan Miao, and Hanwang
  Zhang.
\newblock Learning to segment the tail.
\newblock In {\em CVPR}, 2020.

\bibitem{huynh2021open}
Dat Huynh, Jason Kuen, Zhe Lin, Jiuxiang Gu, and Ehsan Elhamifar.
\newblock Open-vocabulary instance segmentation via robust cross-modal
  pseudo-labeling.
\newblock {\em arXiv:2111.12698}, 2021.

\bibitem{itti1998tpami}
L. Itti, C. Koch, and E. Niebur.
\newblock A model of saliency-based visual attention for rapid scene analysis.
\newblock {\em TPAMI}, 1998.

\bibitem{ji2019iccv}
Xu Ji, Jo{\~a}o~F Henriques, and Andrea Vedaldi.
\newblock Invariant information clustering for unsupervised image
  classification and segmentation.
\newblock In {\em ICCV}, 2019.

\bibitem{judd2009iccv}
Tilke Judd, Krista Ehinger, Frédo Durand, and Antonio Torralba.
\newblock Learning to predict where humans look.
\newblock In {\em ICCV}, 2009.

\bibitem{kamath2021mdetr}
Aishwarya Kamath, Mannat Singh, Yann LeCun, Gabriel Synnaeve, Ishan Misra, and
  Nicolas Carion.
\newblock Mdetr-modulated detection for end-to-end multi-modal understanding.
\newblock In {\em ICCV}, 2021.

\bibitem{kato2019zero}
Naoki Kato, Toshihiko Yamasaki, and Kiyoharu Aizawa.
\newblock Zero-shot semantic segmentation via variational mapping.
\newblock In {\em ICCVW}, 2019.

\bibitem{kuhn1955hungarian}
Harold~W. Kuhn.
\newblock {The Hungarian Method for the Assignment Problem}.
\newblock {\em Naval Research Logistics Quarterly}, 1955.

\bibitem{li2022iclr}
Boyi Li, Kilian~Q Weinberger, Serge Belongie, Vladlen Koltun, and Rene Ranftl.
\newblock Language-driven semantic segmentation.
\newblock In {\em ICLR}, 2022.

\bibitem{li2020consistent}
Peike Li, Yunchao Wei, and Yi Yang.
\newblock Consistent structural relation learning for zero-shot segmentation.
\newblock In {\em NeurIPS}, 2020.

\bibitem{lin2014microsoft}
Tsung-Yi Lin, Michael Maire, Serge Belongie, James Hays, Pietro Perona, Deva
  Ramanan, Piotr Doll{\'a}r, and C~Lawrence Zitnick.
\newblock Microsoft coco: Common objects in context.
\newblock In {\em ECCV}, 2014.

\bibitem{loshchilov2018decoupled}
Ilya Loshchilov and Frank Hutter.
\newblock Decoupled weight decay regularization.
\newblock In {\em ICLR}, 2019.

\bibitem{luddecke2022image}
Timo L{\"u}ddecke and Alexander Ecker.
\newblock Image segmentation using text and image prompts.
\newblock In {\em CVPR}, 2022.

\bibitem{mikolov2013efficient}
Tomas Mikolov, Kai Chen, Greg Corrado, and Jeffrey Dean.
\newblock Efficient estimation of word representations in vector space.
\newblock {\em arXiv:1301.3781}, 2013.

\bibitem{miletari2016vnet}
F. Milletari, N. Navab, and S. Ahmadi.
\newblock V-net: Fully convolutional neural networks for volumetric medical
  image segmentation.
\newblock In {\em 3DV}, 2016.

\bibitem{naseer2021neurips}
Muzammal Naseer, Kanchana Ranasinghe, Salman Khan, Munawar Hayat, Fahad Khan,
  and Ming-Hsuan Yang.
\newblock Intriguing properties of vision transformers.
\newblock In {\em NeurIPS}, 2021.

\bibitem{naseer2021intriguing}
Muhammad~Muzammal Naseer, Kanchana Ranasinghe, Salman~H Khan, Munawar Hayat,
  Fahad Shahbaz~Khan, and Ming-Hsuan Yang.
\newblock Intriguing properties of vision transformers.
\newblock In {\em NeurIPS}, 2021.

\bibitem{ouali2020eccv}
Yassine Ouali, C{\'e}line Hudelot, and Myriam Tami.
\newblock Autoregressive unsupervised image segmentation.
\newblock In {\em ECCV}, 2020.

\bibitem{pastore2021closer}
Giuseppe Pastore, Fabio Cermelli, Yongqin Xian, Massimiliano Mancini, Zeynep
  Akata, and Barbara Caputo.
\newblock A closer look at self-training for zero-label semantic segmentation.
\newblock In {\em CVPR}, 2021.

\bibitem{pennington2014glove}
Jeffrey Pennington, Richard Socher, and Christopher~D Manning.
\newblock Glove: Global vectors for word representation.
\newblock In {\em EMNLP}, 2014.

\bibitem{radford2021icml}
Alec Radford, Jong~Wook Kim, Chris Hallacy, Aditya Ramesh, Gabriel Goh,
  Sandhini Agarwal, Girish Sastry, Amanda Askell, Pamela Mishkin, Jack Clark,
  Gretchen Krueger, and Ilya Sutskever.
\newblock Learning transferable visual models from natural language
  supervision.
\newblock In {\em ICML}, 2021.

\bibitem{rao2021denseclip}
Yongming Rao, Wenliang Zhao, Guangyi Chen, Yansong Tang, Zheng Zhu, Guan Huang,
  Jie Zhou, and Jiwen Lu.
\newblock Denseclip: Language-guided dense prediction with context-aware
  prompting.
\newblock {\em arXiv:2112.01518}, 2021.

\bibitem{russakovsky2015imagenet}
Olga Russakovsky, Jia Deng, Hao Su, Jonathan Krause, Sanjeev Satheesh, Sean Ma,
  Zhiheng Huang, Andrej Karpathy, Aditya Khosla, Michael Bernstein, et~al.
\newblock Imagenet large scale visual recognition challenge.
\newblock {\em IJCV}, 2015.

\bibitem{shin2022unsupervised}
Gyungin Shin, Samuel Albanie, and Weidi Xie.
\newblock Unsupervised salient object detection with spectral cluster voting.
\newblock In {\em CVPRW}, 2022.

\bibitem{shin2022reco}
Gyungin Shin, Weidi Xie, and Samuel Albanie.
\newblock Reco: Retrieve and co-segment for zero-shot transfer.
\newblock In {\em NeurIPS}, 2022.

\bibitem{shin2022namedmask}
Gyungin Shin, Weidi Xie, and Samuel Albanie.
\newblock Namedmask: Distilling segmenters from complementary foundation
  models.
\newblock In {\em CVPRW}, 2023.

\bibitem{simeoni2021localizing}
Oriane Sim\'eoni, Gilles Puy, Huy~V. Vo, Simon Roburin, Spyros Gidaris, Andrei
  Bursuc, Patrick P\'erez, Renaud Marlet, and Jean Ponce.
\newblock Localizing objects with self-supervised transformers and no labels.
\newblock In {\em BMVC}, 2021.

\bibitem{tong2020land}
Xin-Yi Tong, Gui-Song Xia, Qikai Lu, Huanfeng Shen, Shengyang Li, Shucheng You,
  and Liangpei Zhang.
\newblock Land-cover classification with high-resolution remote sensing images
  using transferable deep models.
\newblock {\em Remote Sensing of Environment}, 237:111322, 2020.

\bibitem{van2021iccv}
Wouter Van~Gansbeke, Simon Vandenhende, Stamatios Georgoulis, and Luc Van~Gool.
\newblock Unsupervised semantic segmentation by contrasting object mask
  proposals.
\newblock In {\em ICCV}, 2021.

\bibitem{gansbeke2022maskdistill}
Wouter Van~Gansbeke, Simon Vandenhende, and Luc Van~Gool.
\newblock Discovering object masks with transformers for unsupervised semantic
  segmentation.
\newblock {\em arXiv:2206.06363}, 2022.

\bibitem{vaswani2017attention}
Ashish Vaswani, Noam Shazeer, Niki Parmar, Jakob Uszkoreit, Llion Jones,
  Aidan~N Gomez, {\L}ukasz Kaiser, and Illia Polosukhin.
\newblock Attention is all you need.
\newblock In {\em NeurIPS}, 2017.

\bibitem{vo2022eccv}
Huy~V. Vo, Patrick P{\'e}rez, and Jean Ponce.
\newblock Toward unsupervised, multi-object discovery in large-scale image
  collections.
\newblock In {\em ECCV}, 2020.

\bibitem{vobecky2022arxiv}
Antonin Vobecky, David Hurych, Oriane Siméoni, Spyros Gidaris, Andrei Bursuc,
  Patrick Pérez, and Josef Sivic.
\newblock Drive\&segment: Unsupervised semantic segmentation of urban scenes
  via cross-modal distillation.
\newblock {\em arXiv:2203.11160}, 2022.

\bibitem{WahCUB_200_2011}
C. Wah, S. Branson, P. Welinder, P. Perona, and S. Belongie.
\newblock {The Caltech-UCSD Birds-200-2011 Dataset}.
\newblock Technical report, 2011.

\bibitem{wang2023cut}
Xudong Wang, Rohit Girdhar, Stella~X Yu, and Ishan Misra.
\newblock Cut and learn for unsupervised object detection and instance
  segmentation.
\newblock {\em arXiv:2301.11320}, 2023.

\bibitem{wang2022freesolo}
Xinlong Wang, Zhiding Yu, Shalini De~Mello, Jan Kautz, Anima Anandkumar,
  Chunhua Shen, and Jose~M. Alvarez.
\newblock Freesolo: Learning to segment objects without annotations.
\newblock In {\em CVPR}, 2022.

\bibitem{wang2020DenseCL}
Xinlong Wang, Rufeng Zhang, Chunhua Shen, Tao Kong, and Lei Li.
\newblock Dense contrastive learning for self-supervised visual pre-training.
\newblock In {\em CVPR}, 2021.

\bibitem{wang2022cvpr}
Yangtao Wang, Xi Shen, Shell~Xu Hu, Yuan Yuan, James Crowley, and Dominique
  Vaufreydaz.
\newblock Self-supervised transformers for unsupervised object discovery using
  normalized cut.
\newblock In {\em CVPR}, 2022.

\bibitem{wei2012eccv}
Yichen Wei, Fang Wen, Wangjiang Zhu, and Jian Sun.
\newblock Geodesic saliency using background priors.
\newblock In {\em ECCV}, 2012.

\bibitem{xian2019semantic}
Yongqin Xian, Subhabrata Choudhury, Yang He, Bernt Schiele, and Zeynep Akata.
\newblock Semantic projection network for zero-and few-label semantic
  segmentation.
\newblock In {\em CVPR}, 2019.

\bibitem{xu2021simple}
Mengde Xu, Zheng Zhang, Fangyun Wei, Yutong Lin, Yue Cao, Han Hu, and Xiang
  Bai.
\newblock A simple baseline for zero-shot semantic segmentation with
  pre-trained vision-language model.
\newblock {\em arXiv:2112.14757}, 2021.

\bibitem{zhang2021knet}
Wenwei Zhang, Jiangmiao Pang, Kai Chen, and Chen~Change Loy.
\newblock {K-Net: Towards} unified image segmentation.
\newblock In {\em NeurIPS}, 2021.

\bibitem{zhang2020self}
Xiao Zhang and Michael Maire.
\newblock Self-supervised visual representation learning from hierarchical
  grouping.
\newblock In {\em NeurIPS}, 2020.

\bibitem{zhang2020gicd}
Zhao Zhang, Wenda Jin, Jun Xu, and Ming-Ming Cheng.
\newblock Gradient-induced co-saliency detection.
\newblock In {\em ECCV}, 2020.

\bibitem{zhao2017open}
Hang Zhao, Xavier Puig, Bolei Zhou, Sanja Fidler, and Antonio Torralba.
\newblock Open vocabulary scene parsing.
\newblock In {\em ICCV}, 2017.

\bibitem{zheng2021cvpr}
Ye Zheng, Jiahong Wu, Yongqiang Qin, Faen Zhang, and Li Cui.
\newblock Zero-shot instance segmentation.
\newblock In {\em CVPR}, 2021.

\bibitem{zhou2022maskclip}
Chong Zhou, Chen~Change Loy, and Bo Dai.
\newblock Extract free dense labels from clip.
\newblock In {\em ECCV}, 2022.

\end{thebibliography}
}

\newpage
\appendix
In this supplementary material, we provide a pseudo-code for \methodName (in Sec.~\ref{sec:pseudo-code}) and further details about our experiments (in Sec.~\ref{sec:experiment-details}). Then, we describe additional ablation studies with regards to copy-paste augmentation~\cite{ghiasi2021cvpr} and a hyperparameter choice for temperature used to compute a mask confidence score (in Sec.~\ref{sec:additional-ablation}). Lastly, we show additional visualisations including common failure cases (in Sec.~\ref{sec:additional-visualisations}).

\section{Pseudo-code}\label{sec:pseudo-code}
In~\cref{alg:zesti}, we describe a pseudo-code for a forward pass of \methodName. For readability, we omit operations for bilinear upsampling, which is applied to image features from an image encoder, and non-maximum suppression, which is applied to mask proposals.

\begin{algorithm*}[!htb]
\caption{Pseudo-code for \methodName (using PyTorch-like syntax)}\label{alg:zesti}
\textbf{Input.} a CLIP image encoder $\psi_\mcal{I}^{enc}$, a transformer decoder $\psi_\mcal{I}^{dec}$, a CLIP text encoder $\psi_\mcal{T}$, two feed-forward networks \qcr{FFN}, a projection matrix $\mbf{W} \in \mbb{R}^{e_t \times e_v}$,
an image $x\in \mbb{R}^{3 \times H \times W}$, a set of concepts $\mcal{C}$, queries $Q \in \mbb{R}^{n_q \times e_v}$, threshold $t$,\\ temperature $\tau$\\
\textbf{Output.} predictions for semantic segmentation and instance segmentation
\begin{algorithmic}
\fontfamily{qcr}\selectfont
    \State \comment{extract dense image features}
    \State img\textunderscore feat = $\psi_\mcal{I}^{enc}(x)$ \comment{hxwx$e_v$}\\
    
    \State \comment{extract text features}
    \State text\textunderscore emb = l2\textunderscore normalize($\psi_\mcal{T}(\mcal{C})$, dim=1) \comment{$|\mcal{C}|$x$e_t$}\\

    \State \tikzmk{A}\comment{mask proposals}
    \State $V$ = FFN(img\textunderscore feat.detach()) \comment{hxwx$e_v$}
    \State $Q$ = l2\textunderscore normalize(FFN($\psi_\mcal{I}^{dec}$($Q$, $V$)), dim=1) \comment{$n_q$x$e_v$}
    \State $\mcal{M}$ = sigmoid(mm($Q$, $V$.permute(2, 0, 1))) \comment{$n_q$xhxw}\\
    \tikzmk{B}\boxit{maskproposalscolour}

    \State \tikzmk{A}\comment{inference for semantic segmentation}
    \State \comment{project image features into the text space}
    \State semantic\textunderscore img\textunderscore feat = layer\textunderscore norm(mm($\mbf{W}$, img\textunderscore feat.permute(2, 0, 1))) \comment{$e_t$xhxw}
    \State semantic\textunderscore img\textunderscore feat = l2\textunderscore normalize(semantic\textunderscore img\textunderscore feat, dim=0) 
    \State semantic\textunderscore prediction = argmax(mm(text\textunderscore emb,
    semantic\textunderscore img\textunderscore feat), dim=0) \comment{hxw}\\
    \tikzmk{B}\boxit{semanticcolour}
    
    \State \tikzmk{A}\comment{inference for instance segmentation}
    \State $\mcal{B}$ = $\mcal{M} > t$  \comment{binary masks, $n_q$xhxw}
    \State mask\textunderscore sizes = sum($\mcal{B}$, dim=(1, 2))  \comment{$n_q$}
    \State obj\textunderscore scores = sum($\mcal{B}*\mcal{M}$, dim=(1, 2))/mask\textunderscore sizes  \comment{objectness scores, $n_q$}
    \State avg\textunderscore feat = sum(\\
        \hskip1.0em semantic\textunderscore img\textunderscore  feat.unsqueeze(dim=0)$*\mcal{B}$.unsqueeze(dim=1), dim=(2, 3)\\
        )/mask\textunderscore sizes.unsqueeze(dim=1)  \comment{$n_q$x$e_t$}
    \State avg\textunderscore feat = l2\textunderscore normalize(avg\textunderscore feat, dim=1)

    \State logits = mm(avg\textunderscore feat, text\textunderscore emb.t())  \comment{$n_q$x$|\mcal{C}|$}

    \State conf\textunderscore scores = max(sigmoid(logits$*\tau$), dim=1)$*$obj\textunderscore scores  \comment{confidence scores, $n_q$}
    \State mask\textunderscore classes = argmax(logits, dim=1)  \comment{a category label for each mask, $n_q$}\\
    \tikzmk{B}\boxit{instancecolour}
\end{algorithmic}
mm:\fontfamily{cmr}\selectfont matrix multiplication, $e_v$: a dimension of image features, $e_t$: a dimension of text embeddings,\\
$n_q$: the number of queries
\end{algorithm*}

\section{Experiment details}\label{sec:experiment-details}
Here, we detail the network architectures used for ReCo~\cite{shin2022reco} and NamedMask~\cite{shin2022namedmask} and their differences from \methodName. Next, we describe the details of data augmentations used to train our model.

\subsection{Architecture details for ReCo and NamedMask}
In Sec.~4.3, we compare \methodName to previous methods for unsupervised semantic segmentation with language-image pretraining. 
Here, we describe in more detail the differences between \methodName and ReCo~\cite{shin2022reco}, as well as the concurrent work NamedMask~\cite{shin2022namedmask}.

\mypara{ReCo.} ReCo is composed of two different image encoders and a text encoder. For the former, it adopts DeiT-S/16~\cite{naseer2021neurips} pretrained on Stylised-ImageNet~\cite{geirhos2018iclr} in a supervised manner and ResNet50x16 from CLIP~\cite{radford2021icml}, 
which is used for language-guided co-segmentation~\cite{shin2022reco} together with the text encoder. Unlike ReCo, \methodName involves a single image encoder from CLIP (ViT-B/32 or ViT-B/16) and a corresponding text encoder streamlining the inference process.

\mypara{NamedMask.} NamedMask consists of an image encoder, which is ResNet50~\cite{he2016cvpr} pretrained on the ImageNet1K dataset in an unsupervised manner~\cite{caron2021iccv}, and an image decoder, which is DeepLabv3+~\cite{chen2018eccv}.
It is worth noting that NamedMask follows a conventional semantic segmentation architecture which can only predict a pre-fixed set of classes that the model has seen during training. In contrast, due to the use of a text encoder as a classifier, \methodName can readily predicts categories beyond seen ones after training for a set of concepts.

\subsection{Data augmentation}
For geometric transformations, we use random horizontal flipping with a probability of 0.5, random rescaling with a range of [0.1, 1.0], and random cropping with a size of 384$\times$384. For photometric transformations, we use random colour jittering and gray scaling with a probability of 0.8 and 0.2, respectively. We also use a random gaussian blurring with a kernel size of 10\% of a shorter side of an image.
Lastly, we apply copy-paste augmentations to a set of augmented images so as to synthesise a complex image containing multiple objects.
We set the maximum number of possible objects in an image as 10, which means that we randomly pick from 1 to 10 images and copy-paste an object region of each image represented by its pseudo-mask (obtained by SelfMask~\cite{shin2022unsupervised}).

\section{Additional ablation studies}\label{sec:additional-ablation}
In this section, we conduct further ablation studies about the influence of using copy-paste augmentation and a hyperparameter choice for temperature, which is used to compute a confidence score of a mask proposal. As in the main paper, we use the VOC2012~\cite{Everingham10} \qcr{trainval} split for the ablation studies.

\subsection{Effect of copy-paste}
As mentioned in Sec.~4.1, we apply copy-paste augmentation to synthesise an image with multiple objects following~\cite{shin2022namedmask}.
In~\cref{tab:copy-paste}, we show that it improves performance of \methodName by a large margin with regards to both mIoU (semantic segmentation) and $\maskapfifty$ (instance segmentation).

\begin{figure*}[!t]
    \vspace{-3mm}
    \centering
    \includegraphics[width=0.98\textwidth]{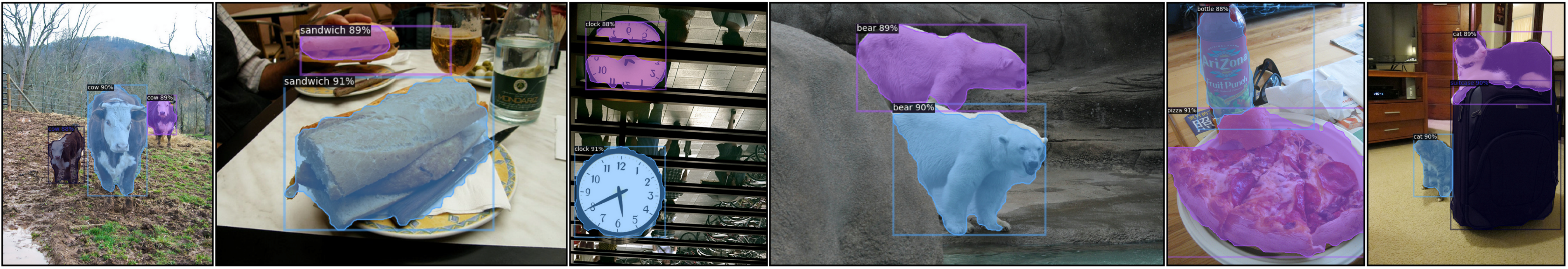}
    \vspace{-3mm}
    \caption{Successful cases of \methodName for instance segmentation on COCO-20K. Confident predictions are shown. Zoom in for details.}
    \label{fig:successful-cases}
    \vspace{-2mm}
\end{figure*}
\begin{figure*}[!th]
    \centering
    \includegraphics[width=0.98\textwidth]{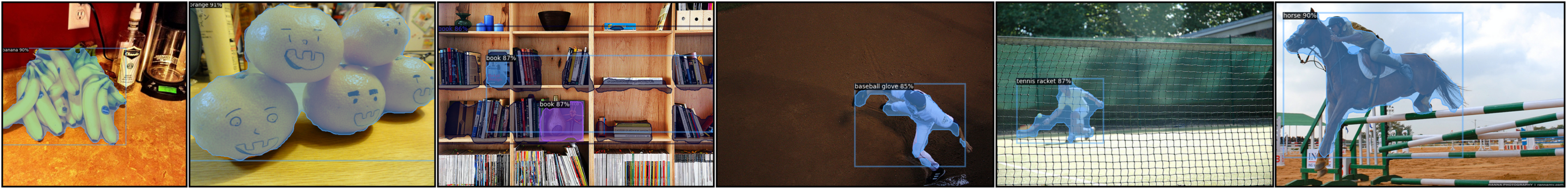}
    \vspace{-3mm}
    \caption{Typical failure cases of \methodName on COCO-20K. (Left half) The model fails to distinguish instances of a same category. (Right half) It struggles to differentiate a category from another 
    which is likely to appear together.}
    \label{fig:failure-cases}
\end{figure*}
\setlength\tabcolsep{0.5pt}
\begin{table}[!htb]
\small
\centering
\begin{booktabs}{c ll}
\toprule
copy-paste & mIoU & $\maskapfifty$\\ \midrule
\xmark & 56.1 & 28.2\\
\checkmark & \textbf{63.7} \green{($+$7.6)} & \textbf{30.9} \green{($+$2.7)} \\ \bottomrule
\end{booktabs}
\vspace{-0.1cm}
\caption{Using copy-paste augmentation allows better performance in terms of both mIoU (for semantic segmentation) and $\maskapfifty$ (for instance segmentaton) on VOC2012 \qcr{trainval}.
}
\label{tab:copy-paste}
\end{table}

\subsection{Effect of temperature}
We compute a confidence score of the mask as a multiplication between the average value of the mask regions and the maximum class probability (see~\cref{alg:zesti}). For the latter, we consider a temperature parameter $\tau$ multiplied to logits for the following sigmoid.
In~\cref{tab:temperature}, we evaluate our model with different values for $\tau$ and observe that setting $\tau$ as 5 yields the best performance. For this reason, we use $\tau=5$ throughout our experiments in the main paper.
\setlength\tabcolsep{6pt}
\begin{table}[!htb]
\centering
\begin{tabular}{c|ccccc}
$\tau$ & 0.1 & 0.5 & 1 & 5 & 10 \\ \thickhline
$\maskapfifty$ & 29.4 & 30.3 & 30.8 & \textbf{30.9} & 30.1  
\end{tabular}
\caption{Effect of temperature $\tau$ on instance segmentation performance of $\methodName$.}
\label{tab:temperature}
\end{table}

\section{Additional visualisations}\label{sec:additional-visualisations}
We visualise additional instance segmentation results of \methodName for successful cases in \cref{fig:successful-cases} and common failure cases in \cref{fig:failure-cases}. We note that our model tends to fail in two situations when (i) an image retrieved for a concept is likely to contain multiple instances for the concept (\eg, bananas), or (ii) a given concept is inclined to appear with another category (\eg, a ``baseball glove'' with a ``person'' wearing it). We conjecture that this is caused by a lack of high purity images (\ie ones that only contain an object of a single category) for each concept in an image index dataset and/or a use of prompt engineering which is not geared towards curating high purity images from the index dataset.

\end{document}